\documentclass[letterpaper]{article} 
\usepackage{aaai25}  
\usepackage{times}  
\usepackage{helvet}  
\usepackage{courier}  
\usepackage[hyphens]{url}  
\usepackage{graphicx} 
\usepackage{natbib}  
\usepackage{caption} 
\usepackage{algorithm}
\usepackage{algorithmic}

\usepackage{multirow}
\usepackage{booktabs}
\usepackage{colortbl}
\usepackage[x11names]{xcolor}
\definecolor{mygray}{gray}{.9}
\usepackage{makecell}
\usepackage{amsmath}
\usepackage{bbding}
\usepackage{cite}

\usepackage{listings}
\DeclareCaptionStyle{ruled}{labelfont=normalfont,labelsep=colon,strut=off} 
\floatstyle{ruled}
\newfloat{listing}{tb}{lst}{}
\floatname{listing}{Listing}
\title{GLCF: A Global-Local Multimodal Coherence Analysis Framework for Talking Face Generation Detection}
\author{
    Xiaocan Chen\textsuperscript{\rm 1},
    Qilin Yin\textsuperscript{\rm 1},
    Jiarui Liu\textsuperscript{\rm 2},
    Wei Lu\textsuperscript{\rm 1}\thanks{Corresponding authors},
    Xiangyang Luo\textsuperscript{\rm 3}\footnotemark[1],
    Jiantao Zhou\textsuperscript{\rm 4}
}
\affiliations{
    \textsuperscript{\rm 1}School of Computer Science and Engineering, MoE Key Laboratory of Information Technology, Guangdong Province Key Laboratory of Information Security Technology, Sun Yat-sen University, Guangzhou 510006, China\\
    \textsuperscript{\rm 2}Alibaba Group, Hangzhou, China\\
    \textsuperscript{\rm 3}State Key Laboratory of Mathematical Engineering and Advanced Computing, Zhengzhou 450002, China\\
    \textsuperscript{\rm 4}State Key Laboratory of Internet of Things for Smart City, Department of Computer and Information Science, Faculty of Science and Technology, University of Macau, Macau 999078, China\\
    chenxc33@mail2.sysu.edu.cn, yinqlin@mail2.sysu.edu.cn, liujiarui.ljr@alibaba-inc.com, luwei3@mail.sysu.edu.cn, luoxy$\_$ieu@sina.com, jtzhou@um.edu.mo
}

\begin{document}

\maketitle

\begin{abstract}
Talking face generation (TFG) allows for producing lifelike talking videos of any character using only facial images and accompanying text.
Abuse of this technology could pose significant risks to society, creating the urgent need for research into corresponding detection methods.
However, research in this field has been hindered by the lack of public datasets.
In this paper, we construct the first large-scale multi-scenario talking face dataset (MSTF), which contains 22 audio and video forgery techniques, filling the gap of datasets in this field. The dataset covers 11 generation scenarios and more than 20 semantic scenarios, closer to the practical application scenario of TFG. 
Besides, we also propose a TFG detection framework, which leverages the analysis of both global and local coherence in the multimodal content of TFG videos. Therefore, a region-focused smoothness detection module (RSFDM) and a discrepancy capture-time frame aggregation module (DCTAM) are introduced to evaluate the global temporal coherence of TFG videos, aggregating multi-grained spatial information. Additionally, a visual-audio fusion module (V-AFM) is designed to evaluate audiovisual coherence within a localized temporal perspective.
Comprehensive experiments demonstrate the reasonableness and challenges of our datasets, while also indicating the superiority of our proposed method compared to the state-of-the-art deepfake detection approaches.
\end{abstract}

\section{Introduction}
Recently, generative AI technology has achieved some significant advancements.  Notably, digital human generation technology, a real-world application of AI generation technology, has been widely used in commercial live broadcasting and other fields, fostering societal and economic progress.

Talking face generation (TFG) represents a pivotal technology in creating digital humans, inherently differing from traditional deepfake approaches. While traditional deepfake methods rely on substituting facial regions within videos to replace identities, the quality of the generated video is constrained by the existing video content. Through talking face generation technology, we can create a highly realistic speech video of a specific individual with just a single image of that person and a textual script, or alternatively, a video clip and an audio segment. The TFG videos demonstrate superior visual quality, exhibiting more realistic details, such as illumination uniformity, lip movements, and so on. Besides, this technology fundamentally advances the forgery process, transitioning from a reliance on strong references to mere weak ones, diminishing the difficulty of forgery.
Thus, someone can effortlessly forge videos of renowned politicians making inappropriate statements through this technology, triggering public panic and a crisis of confidence.

\begin{figure}[t]
	\centering
	{\includegraphics[width=3.1in]{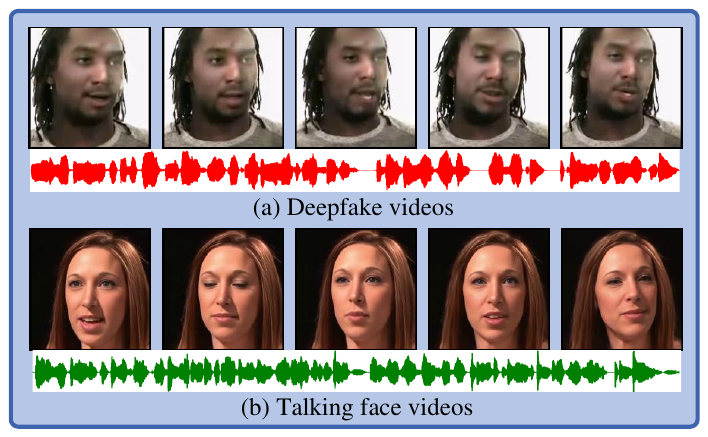}}
	\caption{(a) Deepfake videos. (b) Talking face videos.\\
 Compared to deepfake videos, talking face videos demonstrate superior visual quality, exhibiting more realistic details, such as illumination uniformity, lip movements, teeth and so on.}
	\label{show}
\end{figure}
\arrayrulecolor{black}
\begin{table*}[t]
	\begin{center}
		\begin{tabular}{lccccccc}
			\Xhline{1.2pt}
			\rowcolor{gray!20}
			\textbf{DataSet} \multirow{2}{*} & \textbf{Date} \multirow{2}{*} & \textbf{Modality} \multirow{2}{*}& \textbf{Real} & \textbf{Fake} & \textbf{Total} & \textbf{Manipulation}  & \textbf{Audio Label} \\
			\rowcolor{gray!20}
			& & & \textbf{Video/Audio} & \textbf{Video/Audio} & \textbf{Video/Audio} & \textbf{Method} &  \\
			
			\hline 
			\hline
			DF-TIMIT & 2018 & V & 320 & 640 & 960 & 2 & No \\ 
			FF++ & 2019 & V & 1,000 & 4,000 & 5,000 & 4 & No \\ 
			DeeperForensics & 2020 & V & 360 & 3,068 & 3,431 & 5 & No \\
			FFIW & 2021 & V & 10,000 & 10,000 & 20,000 & 1 & No \\
			KoDF & 2021 & V & 62,166 & 175,776 & 237,942 & 5 & No \\
			DF\_Platter & 2023 & V & 133,260 & 132,496 & 265,756 & 3 & No \\
			\hline
			ASVspoof2019 & 2019 & A & 10,256 & 90,192 & 100,448 & 19 & Yes \\
			WaveFake & 2021 & A & 0 & 5,160 & 5,160 & 7 & Yes \\
			CFAD & 2023 & A & 38,600 & 77,200 & 115,800 & 12 & Yes \\
			\hline
			DFDC & 2020 & A/V & 23,654 & 104,500 & 128,154 & 8 & No\\
			FakeAVCeleb & 2021 & A/V & 500 & 19,500 & 20,000 & 4 & Yes \\
			LAV-DF & 2022 & A/V & 36,431 & 99,873 & 136,304 & 2 & Yes \\
			DefakeAVMiT & 2023 & A/V & 540 & 6,480 & 7,020 & 5 & Yes\\
			\hline
			MSTF(Ours) & 2024 & A/V & 37,059 & 106,695 & 143,754 & 22 & Yes \\
			\Xhline{1.2pt}
		\end{tabular}
	\end{center}
\caption{Quantitative comparison of MSTF with existing publicly available deepfake datasets.}
\end{table*}

For TFG detection, existing deepfake detection methods exhibit limited transferability.
\cite{zhao2021multi} and \cite{peng2024deepfakes} capture pixel-level forgery traces from the spatial domain. Since the superior visual quality of TFG videos, it is difficult to extract forgery traces solely from the spatial domain.
As talking face generation relies solely on a single weak reference, the frequency domain characteristics of the videos do not exhibit the significant anomalies observed in deepfake videos. Thus, frequency-focused methods, such as \cite{mejri2021leveraging} and \cite{tan2024frequency}, are difficult to attain robust performance. 
Since TFG offers more precise audio-visual synchronization, the existing detection techniques utilize multi-modal audiovisual interaction for deepfake video detection, such as \cite{zhou2021joint} and \cite{feng2023self}, face significant limitations when applied to talking face videos.
Works from other fields, such as \cite{yang2024robust} and \cite{pang2024heterogeneous}, are insightful, but can't be directly applied.

Therefore, developing targeted detection methods for talking face generation is critical.
Currently, the public datasets are mainly visual unimodality. Multi-modal datasets emphasize face replacement forgery, making it challenging to support research on talking face detection.
To fill this gap, we propose the first challenging and multi-scenario talking face dataset (MSTF) with the following characteristics: (1) Large-scale, comprising over 100,000 entries. (2) Containing 22 forgery techniques, which is the largest number of forgery techniques among the current mainstream datasets. (3) 11 kinds of generation scenarios. (4) More than 20 kinds of semantic scenarios. Thus, MSTF could effectively support the research of the corresponding detection methods. 

Besides, we also propose a framework to analyze both global and local coherence in the multimodal content of the videos, achieving high-precision TFG detection.
Given the superior visual quality in TFG videos, we focus on the global temporal coherence across the frames to expose more elusive forgery artifacts.
We design a region-focused smoothness detection module (RSFDM), utilizing an attention-like mechanism to target the motion areas and extract the motion information, capturing the incoherent forgery traces during frame transitions.
From the local temporal perspective, a visual-audio fusion module (V-AFM) is designed to evaluate the modality coherence which is the primary strength of TFG compared to deepfake but also the aspect most susceptible to errors. 
Since frames in TFG videos are produced sequentially, they inevitably exhibit incoherence in the spatial domain, which may be regarded as motion areas in RSFDM. Given the highly precise audiovisual synchronization in TFG, it is difficult to comprehensively capture subtle audiovisual coherence solely through V-AFM. Thus, we design a discrepancy capture-time frame aggregation module (DCTAM) to capture subtle differences between frames, complementing the function of RSFDM. Additionally, it improves the precision of modality alignment, thereby enhancing the performance of V-AFM. Firstly it quantifies the magnitude of pixel differences, activating anomalous regions through discrepancy metrics, and then it adaptively integrates spatial information from adjacent frames at multiple granularities. This module serves as a link between RSFDM and V-AFM to make our framework more cohesive.

We conduct a series of experiments, demonstrating the challenges and advancement of
our dataset. Compared to the current state-of-the-art deepfake detection methods, our framework achieves the best performance on the talking face dataset, while also exhibiting good performance on other deepfake datasets.
The main contributions of this work are summarized as follows:
\begin{itemize}
	\item[$\bullet$] We propose a large-scale multi-scenario talking face dataset, which contains 22 audio and video forgery methods and 11 generation scenarios.
	\item[$\bullet$] We also propose a TFG detection framework that analyzes multi-granularity global spatiotemporal coherence and multimodal local coherence.
	\item[$\bullet$] An RSFDM is proposed to target the motion areas, capturing the incoherent traces during global frame transitions. A V-AFM is introduced to evaluate the audiovisual modality incoherence from a local temporal perspective.
    \item[$\bullet$] A DCTAM is designed to quantify and activate anomalous pixel regions, and then adaptively aggregates spatial information from adjacent frames at multiple granularities, serving as a link between RSFDM and V-AFM.
\end{itemize}

\section{Talking Face Dataset}
\subsection{Comparision with Existing Datasets}
Constrained by the immaturity of forgery techniques and limited computational resources during that period, deepfake datasets such as DF-TIMIT \cite{korshunov2018deepfakes}, DF \cite{jiang2020deeperforensics} and FF++ \cite{rossler2019faceforensics++} are typically characterized by their small scale and noticeable visual artifacts.

With the advancement in computational power and the technical improvements, higher-quality and larger-scale deepfake datasets, such as FFIW \cite{zhou2021face}, KoDF \cite{kwon2021kodf} and DF\_Platter \cite{narayan2023df} have emerged in succession. However, these datasets primarily focus on video-only unimodal forgery. Subsequently, with the maturation of audio forgery technology, datasets such as ASV2019 \cite{todisco2019asvspoof}, WaveFake \cite{frank2021wavefake}, and CFAD \cite{ma2022cfad}, which focus solely on the audio modality, have been proposed, further diversifying the forms of forgery data.

DFDC \cite{dolhansky2020deepfake}, the first large-scale dataset with audio-video forgery, mostly uses GAN-based face swapping but lacks clear audio authenticity labels, limiting its application potential.
FakeAVCeleb \cite{khalid2021fakeavceleb}, introduced in 2021, holds significant importance for multimodal deepfake detection.
However, FakeAVCeleb and DefakeAVMiT \cite{yang2023avoid} are constrained by limited talking face generation methods and video quantity, hindering their coverage of complex generated scenarios. LAV-DF \cite{cai2023glitch} is proposed to support the novel research task of multi-modal forgery temporal localization.

Thus, these datasets could not meet the need for research on talking face detection methods.
To fill this gap, we introduce a large-scale, multi-scenario talking face dataset (MSTF) that includes five generation methods and seventeen audio forgery techniques. Though combining various reference inputs, we simulate 11 generation scenarios. These generation scenarios ensure that the TFG videos exhibit multi-level coherence across different modalities, including facial expressions, eye movements, and so on. Additionally, images from different shooting scenes and audio on various topics are combined to make our dataset encompass nearly 40 semantic scenarios. The diverse scenarios make our dataset more relevant to real-world applications of TFG technique, providing a foundation for the research of robust detection methods.

\subsection{Collection}
To better cover a variety of talking face generation scenarios, we collected multiple image, audio, and video datasets.
\subsubsection{Image and Video Collection}
We collect and utilize CelebA-HQ \cite{karras2018progressive}, Cream-D \cite{cao2014crema}, VoxCeleb \cite{nagrani2017voxceleb}, VoxCeleb2 \cite{chung2018voxceleb2}, Mead \cite{wang2020mead}, DFDC and LSR2 \cite{son2017lip} as source data. We also collect speech videos from YouTube with some high-quality videos, including scenes from news broadcasts and interviews.

\subsubsection{Audio Collection}
We additionally collect some speech datasets to enrich our speaker usage scenarios.
We extract all the forged audio from FakeAVCeleb. Besides, we collect LibriSpeech \cite{panayotov2015librispeech} and ASVspoof2019 as driven-audio.

\begin{figure}[t]
	\centering
	{\includegraphics[width=3.3in]{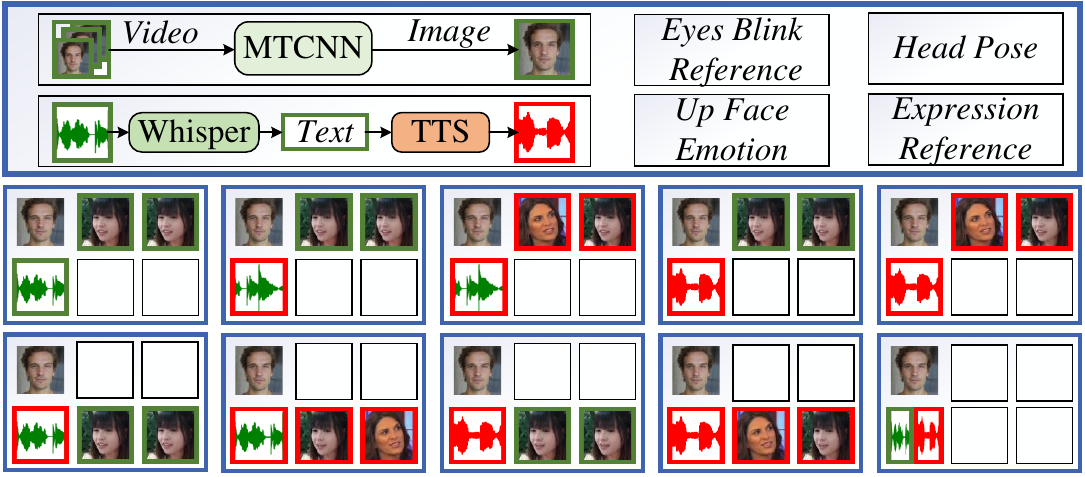}}
	\caption{The construction of 11 scenarios. The six boxes in the above big image correspond to 
the six boxes in each small image below. The green box indicates inputs from the same source, while the red box is the opposite. Green audio denotes genuine audio, whereas red audio signifies forgery audio. Blank means no input.}
	\label{constrution}
\end{figure}
\subsection{Dataset Construction}
\subsubsection{FakeVox} To simulate the text-guided audio synthesis process in real-world scenarios, we generate additional fake audio using text-to-speech (TTS) technology provided by Tencent. Whisper (Radford et al. 2023) is used to convert the audio data to text. English texts with sentences between 85 and 160 words in length are selected from VoxCeleb. Finally, we respectively generate an emotional and a standard audio datasets, each containing 2000 samples.

\begin{figure*}[htbp]
	\centering
	{\includegraphics[width=6.3in]{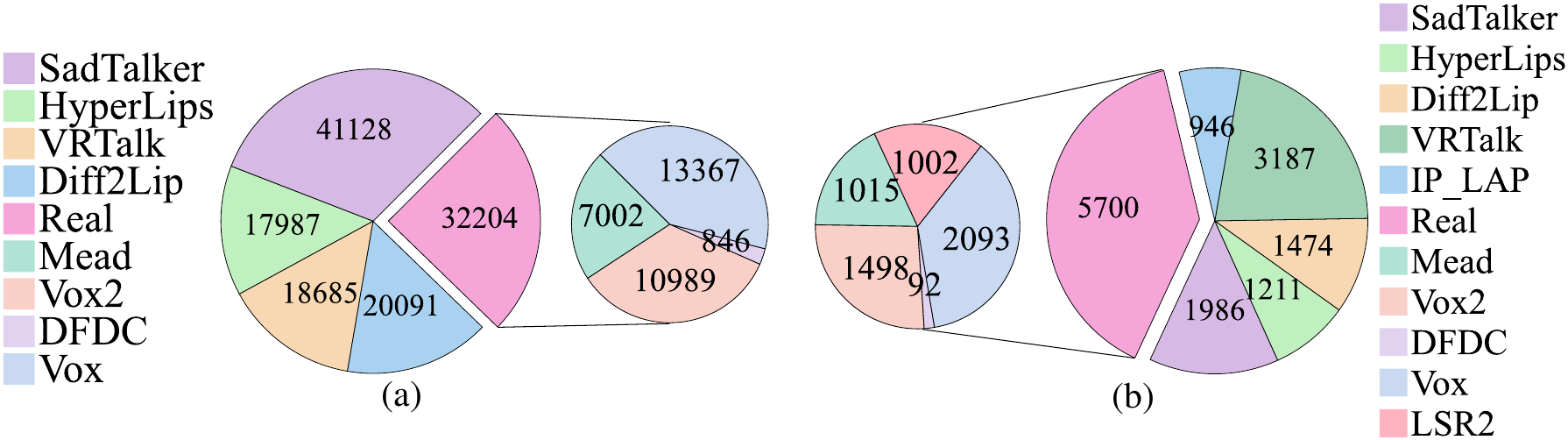}}
	\caption{(a) The distribution of training set. (b) The distribution of testing set.}
     \label{pie}
\end{figure*}

\subsubsection{Generation Scenario}
As shown in Figure \ref{constrution}, the green box indicates inputs from the same source, while the red box indicates inputs from different videos.
TFG technology requires at least one audio segment, and an image or a video to generate a video.
Therefore, we firstly construct two fundamental generation scenarios: genuine audio-driven generation and the forgery one, which are the most common due to their low cost and widespread application, shown in the last small image of Figure \ref{constrution}.
Furthermore, aiming to create more realistic videos that contain richer body language, we introduce additional references such as expressions, head movements, and so on. To generate videos that exhibit optimal coherence between visual and auditory semantics, we design a scenario in which the references and the audio originate from the same video.
However, the above scenarios do not encompass all complex application scenarios.
Thus, another kind of scenario is designed, which is relatively easy to implement and closer to real-world applications. The audio source for this scene is different from the reference video, but all the reference videos are the same one.
Finally, we design a kind of scenario where all inputs are completely different, further enriching our dataset scenarios.
To achieve the large-scale generation of these not reference scenarios, We collect and utilize HyperLips \cite{chen2023hyperlips}, Diff2Lip \cite{mukhopadhyay2024diff2lip} and IP\_LAP \cite{zhong2023identity}. SadTalker \cite{zhang2023sadtalker} is selected for referenced generation, which includes eyes blinking and head pose. As shown in Figure \ref{constrution}, it is designed with 5 scenarios in the first row of small charts. VideoRetalking \cite{cheng2022videoretalking}, which supports inputting references of upper face emotion and expression, is chosen to construct 4 scenarios, shown in the first four subgraphs displayed in the last row. 

\subsection{Dataset Distribution}
In figure \ref{pie}, we present the generation distribution of the dataset on each generation method.
To simulate the detection of potential novel real-world scenarios, we add a small portion of the LSR2 dataset to our testing set, which remains unexposed to the training phase, thereby increasing the complexity of the presented scenes. 
Given the rapid advancements in generative artificial intelligence and talking face generation technologies, we incorporate the IP\_ LAP, which also remains unexposed to the training set, to generate a small portion of testing cases. This cross-domain evaluation emphasizes the necessity for detection methods to exhibit generalization abilities to ensure their superior performance within the testing set. Hence, these configurations facilitate the exploration of generalized, adaptable detection methods, enabling us to effectively tackle unknown and sophisticated forgery challenges.

Furthermore, we combine videos from different shooting scenes and audios on different topics to simulate more than 20 semantic scenarios, resulting in a total of about 40 different scenarios when combined with generation scenarios.

\section{Methodology}
\subsection{Overview}
\begin{figure*}[t]
	\centering
    {\includegraphics[width=5.9in]{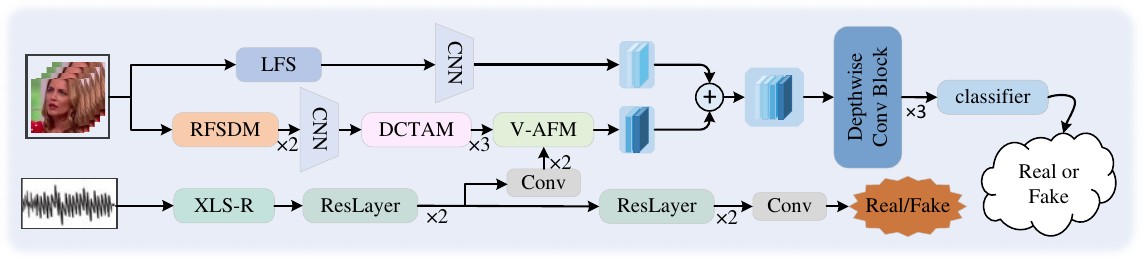}}
	\caption{The overall framework of our proposed GLCF. RFSDM and DCTAM extract spatial features, interacting with audio features in V-AFM to get fusion information which is concatenated with frequency domain features extracted by the LFS.}
	\label{base}
	\centering
	{\includegraphics[width=6.8in]{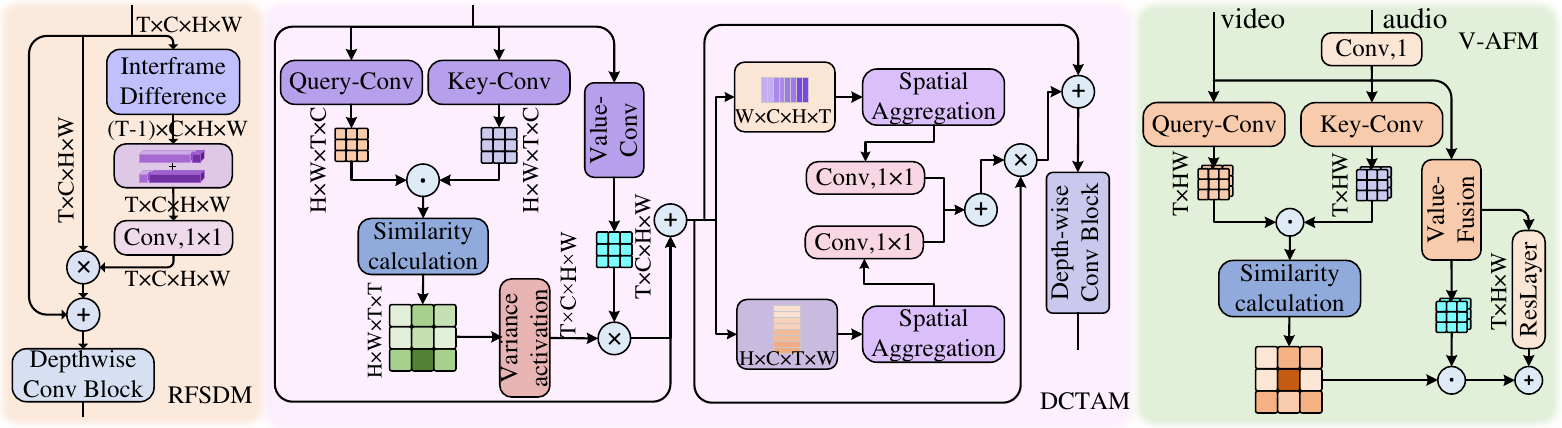}}
	\caption{The detail of RFSDM, DCTAM, V-AFM. \(\oplus\) and \(\otimes\) stand for element-wise addition and multiplication.}
	\label{detailm}
\end{figure*}

As shown in Figure \ref{base}, we propose a global-local multimodal coherence analysis framework for talking face generation detection. 
Our framework includes two video streams and one audio stream. Frequency features are extracted from the first video stream. The second one is processed through two global visual coherence detection modules, mapped into a high-dimensional feature vector that contains abundant information on anomalous regions. Processed through wav2vec and residual layers, the audio stream interactive fusions with the information of second video streaming. The fusion information and frequency features are concatenated to further extraction of deep features.

We find that talking face videos often exhibit incoherence during transitions between frames.
To address this, a \textbf{Region-Focused Smoothness Detection Module} is designed to focus on the principal regions of facial movements at a coarse-grained level, extracting motion information between inter-frames. It employs an attention-like mechanism from a global temporal perspective, effectively capturing the incoherence of frame transitions.

\begin{figure}[t]
	\centering
	{\includegraphics[width=3in]{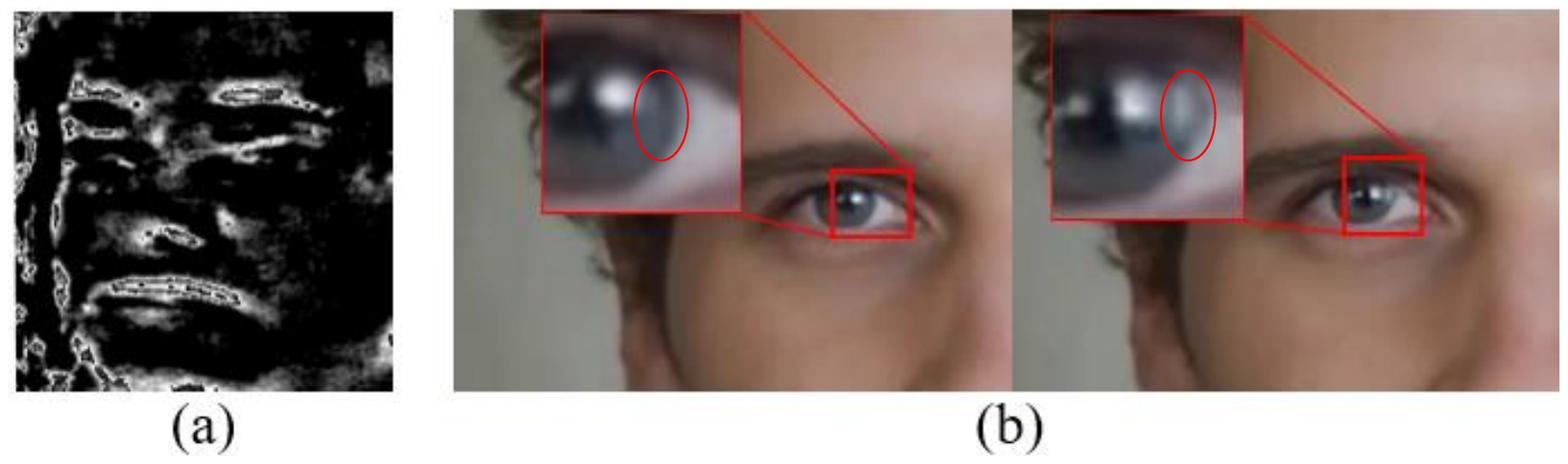}}
	\caption{(a) The diagram of the difference between frames. (b) The slight discrepancy in eyeball shape between frames.}
	\label{difference}
\end{figure}

Frame-by-frame generation methodologies inevitably result in incoherence to the global temporal frames. As shown in Figure \ref{difference} (b), the same region in a generation video exhibits incoherence across different frames. This slight discrepancy may be interpreted as a motion area by RFSDM, which is an undesired outcome. Through the \textbf{Discrepancy Capture-Time Frame Aggregation Module}, we quantify pixel-level spatial discrepancies and propose a differential assessment metric to activate anomalous regions, thereby capturing these subtle inter-frame differences at a fine-grained level. Audiovisual coherence analysis is an effective method for TFG detection. Given the differences in feature extraction between audio and visual data, direct modality interaction often fails to achieve precise temporal alignment, making it inadequate for TFG detection. Therefore, through DCTAM, the spatial information from local temporal frames is aggregated across the multiple granularities of receptive fields.

Besides, the raw audio is transformed into feature vectors through a pre-trained wav2vec model and residual network layers.
Then it is input along with the visual information into an innovative \textbf{Visual-Audio Fusion Module}, which is introduced to compute similarity metrics between the audiovisual modalities from a local temporal perspective.
This module facilitates the detection of inter-modal incoherence, which encompasses not only the generation flaws of lip movement but also other aspects like head pose.

Spatial analysis is easily disturbed by noise and compression. In contrast, the frequency information can reveal the deep structural characteristics of the image. Based on this, we refer to the \textbf{Local Frequency Statistic Method} (LFS) proposed by Qian et al. \cite{qian2020thinking}. By referring to this module, we successfully extract the statistical frequency domain distribution information, which is concatenated with the audiovisual information for the subsequent depthwise convolution block.

\subsection{Region-Focused Smoothness Detection Module}
Calculating the inter-frame difference is an effective way of reflecting the motion information of video objects. The input frame is denoted as $I \in R^{C\times H\times W}$, $C$ is the number of channels, $H$ and $W$ are spatial dimensions. As input $T$ frames $[I_{1}... I_{T}]$ , We got input feature $F \in R^{F\times C\times H\times W}$. We calculate the difference between the two leading frames to get the $[D_{1}... D_{T-1}]$, where $ D_{i-1} = I_{i}-I_{i-1}$ .Then we respectively copy the first and last one of $[D_{1}... D_{T-1}]$ to get $[D_{1}, D_{1}, D_{2}... D_{T-1}]$ and $[D_{1}, D{2}... D_{T-1}, D_{T-1}]$. By adding them, we get $\hat{D} = [\hat{D}_{1}...\hat{D}_{T}]$, which denotes the difference between the corresponding frame and the two preceding and following frames. Finally, We use 2D CNN to adaptively learn the weights of the channels, fusing the inter-frame difference information into the original input as regions of interest of the image pixels. This Module can be formulated as:
\begin{equation}
	F_{out} = F_{in} + F_{in} \cdot Conv_{1\times 1}(\hat{F}).
\end{equation}
This module not only enables the feature maps to retain inter-frame transition coherence information but also reflects the motion regions in videos due to the frame difference itself, drawing the model's attention to the main forged regions such as the mouth. Subsequently, the output is fed into a depth-wise block for further extraction of deep features.

\subsection{Discrepancy Capture-Time Frame Aggregation Module}
Despite its simplicity and efficacy, the inter-frame differential computation is inadequate for discerning subtle forgery indications between frames. As illustrated in Figure \ref{difference}, these inconspicuous forgeries may be disregarded or misidentified as motion-related regions or mere motion information, resulting in insufficient attention being paid to them. To address this limitation, we devise the Discrepancy Capture-Time Frame Aggregation Module, which could adaptively capture these subtle differences by the modified attention mechanism.

We transform the input feature $F \in R^{F\times C\times H\times W}$ into the matrix $F_{Q}, F_{K}, F_{V}$ by 2D CNN. We calculate the similarity of $F_{Q}$ and $F_{K}$ to obtain the Attention Score Matrix $A \in R^{H\times W\times T\times T}$, which represents the similarity of a pixel point between a particular frame and all frames. It should be noted that the subtle disparities that arise during the frame-by-frame generation process are inherently stochastic. In particular, upon examination of pixels within a specified region, the value of similarity between abnormal frames and normal frames will be much smaller than the one between normal frames. Consequently, we calculate the variance in the $T$ dimension of the attention score matrix, detecting the aforementioned phenomenon. For the matrix $A$, we perform a summation operation on its last dimension to get $A \in R^{H\times W\times T}$. Then, the variance is calculated along dimension $T$ to generate variance matrix $V \in R^{1\times 1\times H\times W}$.
We pick the k-th smallest variance to be the threshold of activation, greater than the value masked as 1, less than it is set to 0. This operation is similar to the activation layer, except that we get the threshold of activation by calculating the variance, so we call it the variance-activated layer, which is better than activation directly using the ReLU function. This process could be written as:
\begin{equation} 
	F = F + \alpha \cdot F_{V} \cdot \Phi(F_{Q}\times F_{K}),
\end{equation}
where $\Phi$ represents the variance activation function, and $\alpha$ is an adaptive weight.
One of the challenges associated with multimodal interaction is the precise alignment of audiovisual features at a temporal granularity. Thus, we aggregate temporal information from multiple granularity pairs.
As shown in Figure \ref{detailm}, we reshape $F$ into horizontal and vertical feature vectors $F_{v} \in R^{W\times C\times H\times T}, F_{h} \in R^{H\times C\times T\times W} $, which can retain more detailed visual information. After that, we perform downsampling and convolution of different sizes on the features $F_{v}$ and $F_{h}$, and finally perform upsampling and fusion. This process can be described as:
\begin{equation}
	\begin{split}
    \begin{aligned}
	F_{v} = & \frac{1}{3} \cdot (Conv_{1\times3}F_{v} + Up_{1\times2}(Conv_{1\times3}(Down_{1\times2} \\ & (F_{v}))) + Up_{1\times4}(Conv_{1\times3}(Down_{1\times4}(F_{v})))).
    \end{aligned}
	\end{split}
\end{equation}
The handling of $F_h$ is similar, except that the convolution kernel is transposed.

\subsection{Audio Stream and V-AFM}
\subsubsection{Audio Stream}
We use a pre-trained wav2vec model and a two-layer residual network to map the input audio stream into the feature space, obtaining $F_{a}$.
The output will be fed into two streams. One of them introduces a complete forgery audio detection stream, which directly predicts the authenticity of the audio. This stream also facilitates the addition of the audio loss function $L_{a}$, better fine-tuning of Wav2Vec and training of ResLayer.
The total loss function is:
\begin{equation}
	L = L_{va} + L_{a}.
\end{equation} 
Appropriate intermediate features are conducive to modal fusion. We input another output stream into two $3\times3$ CNNs to obtain intermediate audio feature vectors. These vectors are input into the audio-visual coherence detection module.

\subsubsection{Video-Audio Fusion Module}
We design an audiovisual fusion module based on a multi-head cross-attention mechanism, the main purpose of which is to detect the coherence between audiovisual modes. Specifically, We input audio features into a 1D CNN to align the feature dimension with the visual feature vector, and then further input a 3D CNN to obtain $Query \in R^{C\times\ T\times H\times W}$. Similarly, visual features are entered into the 3D CNN to get the $Key \in R^{C\times\ T\times H\times W}$. $Value$ is a linear combination of audiovisual features, and the entire calculation process can be expressed as:
\begin{equation}
	F_{vai} = V_{i}\times \mathrm{softmax}(\frac{Q_{i}K_{i}^{T}}{H}),
\end{equation}
\begin{equation}
	F_{va} = \Psi(V) + Conv_{1\times 1}(\mathrm{Concat}(F_{vai})),
\end{equation}
where $\Psi$ represents the Residual layer, $i$ represents the head.
	
\section{Experiments}
\subsection{Experimental Settings}
\subsubsection{Datasets} To evaluate our dataset and methodology, we conduct experiments on our MSTF and three other challenging datasets: FaceForensis++, FakeAVCeleb, and DFDC.
\begin{itemize}
	\item[$\bullet$] MSTF is our large-scale talking face dataset, comprising 130,095 training videos and 14,504 testing videos,
	\item[$\bullet$] FaceForensics++ is a conventional deepfake dataset. We utilized its visually blurred low-quality (LQ) version to contrast and evaluate the challenges posed by MSTF. Given the incompatibility of multimodal methods on this dataset, we exclusively employ unimodal forgery detection methods for our experiments. 
	\item[$\bullet$] DFDC incorporates advanced face-swapping forgeries, posing a significant challenge for detection. We use 4080 fake videos and 1133 real videos for the experiment.
	\item[$\bullet$] FakeAVCeleb, a popular deepfake multimodal dataset, contains 500 real and 20,000 forged videos for evaluating forgery detection methods.
\end{itemize}  

\subsubsection{Evaluation metrics}
In our experiments, we use the accuracy rate (ACC) for evaluation.


\subsubsection{Implementation Details} 
We use MTCNN to detect faces in all datasets and then crop them. We adjust the face images to 256*256 to retain more facial information. We use NVIDIA A100 Tensor Core GPU.

\subsection{Performance Comparisons}
\subsubsection{Intra-Dataset Comparisons}
We select several frame-based detection methods for comparison, including F3Net \cite{qian2020thinking}, RFM \cite{wang2021representative}, Multi-att \cite{zhao2021multi}, RECCE \cite{cao2022end}, MiNet \cite{ba2024exposing}. Additionally, we incorporate several multimodal video detection approaches, including Joint A-V \cite{zhou2021joint}, AVSSD \cite{sung2023hearing}, and AVD2-DWF \cite{wang2024avt2}. It is noteworthy that the code for ReCCE, Joint A-V, and AVD2-DWF are not fully open-sourced, thus we try our best to replicate it based on the original documentation. All these methods have been published in reputable journals or conferences. The experimental results are presented in Table 2. 

Experimental results show that our raw MSTF dataset is even harder to detect content from than the well-known and challenging FF++ dataset(C40), revealing the necessity of specialized detection for talking face generation and the effectiveness and validity of the MSTF dataset in facilitating such endeavors.
\begin{table}[t]
\begin{center}
	\begin{tabular}{l|c|c|c|c}
		\Xhline{1.2pt}
		Methods         & FF++    & FakeAV & DFDC    & MSTF    \\
		\hline
		F3Net           & \textbf{0.8767} & 0.9829  & \underline{0.9245} & 0.8516 \\
		RFM             & 0.8410 & 0.9634    & 0.6406 & 0.8521 \\
		Multi-att       & 0.8740 & \underline{0.9833}    & 0.8976 & 0.7871 \\
		RECCE           & \underline{0.8742} & 0.9831    & 0.9125 & 0.8191 \\
		MiNet           & 0.8626 & 0.9768    & 0.8933 & \underline{0.8612} \\
		Joint A-V       & --      & 0.9769    & 0.7773 & 0.8454 \\
		AVSSD           & --      & 0.9760    & 0.6445 & 0.8292 \\
		AVD2-DWF        & --      & 0.9736    & 0.6758 & 0.7821 \\
		\hline
		Ours            & --      & \textbf{0.9892}     & \textbf{0.9258} & \textbf{0.8849} \\
		\Xhline{1.2pt}
	\end{tabular}
\end{center}
\caption{Comparison result on FF++(C40), FakeAVCeleb, DFDC and MSTF in terms of a accuracy}
\end{table}
In the meantime, our method achieves the best performance on the FakeAVCeleb, DFDC, and MSTF datasets. On the MSTF dataset, it surpasses the second-best performer, MiNet, by 2.37\%. For the FakeAVCeleb dataset, focusing on the binary classification of genuine and fake samples, our method also outperforms the second-best performer, Multi-att, by 0.59\%. Similarly, on DFDC, we outperform F3Net, the second-best performer, by a narrow margin of 0.13\%. 
Experimental results demonstrate that our methodology is not merely confined to the specific domain of talking face generation detection but also exhibits a significant degree of generalizability, enabling it to adeptly address a wide array of forgery tactics, including traditional deepfake.

\begin{table}[t]
\begin{center}
	\begin{tabular}{l|c|c}
		\Xhline{1.2pt}
		Methods         & FakeAVCeleb & MSTF    \\
		\hline
		F3Net           & 0.9829 & 0.5611 \\
		RFM             & 0.9634    & 0.5783 \\
		Multi-att       & \underline{0.9833}    & 0.5148 \\
		RECCE           & 0.9831    & 0.5941 \\
		MiNet           & 0.9768    & 0.5438 \\
		Joint A-V       & 0.9769    & 0.4124 \\
		AVSSD           & 0.9760    & 0.5981 \\
		AVD2-DWF        & 0.9736    & \underline{0.6000} \\
		\hline
		Ours            & \textbf{0.9892}   & \textbf{0.6194} \\
		\Xhline{1.2pt}
	\end{tabular}
\end{center}
\caption{Comparison result on cross-dataset generalization in terms of accuracy.}
\end{table}
\subsubsection{Inter-Dataset Comparisons}
We perform training on the FakeAVCeleb and perform testing on FakeAVCeleb and our proposed MSTF. The experimental results are shown in Table 3. Our method simultaneously achieves the best performance on both datasets. This indicates that our method captures the common forgery traces between traditional deepfake and talking face videos, rather than relying on overfitting. 

Overall, methods trained on deepfake datasets do not exhibit good generalization on the MSTF, further highlighting the importance of dedicated datasets of talking face videos.

\subsection{Ablation Study}
Given our method's focus on talking face generation, we conduct a series of ablation studies on the MSTF dataset to evaluate the importance of our four primary modules.  As shown in Table 4, our innovative RFSDM, DCTAM, and V-AFM modules each contributed to an increase in model performance by at least 3\%, with the V-AFM module, specifically designed for modal consistency detection, demonstrating an even more substantial improvement of 8.27\%. Although the LFS module has the least impact on model performance, it indicates that feature completion in the frequency domain is beneficial for detecting talking face forgery. Achieving better results may require further design for TFG.

\begin{table}[t]
	\begin{center}
		\begin{tabular}{ccccc}
			\Xhline{1.2pt}
			\multicolumn{4}{c}{Method} &   MSTF \\
			LFS & RFSDM & DCTAM & V-AFM  &   \\ 
			\hline 
			--- & \Checkmark & \Checkmark & \Checkmark & \underline{0.8625} \\
			\Checkmark & --- & \Checkmark & \Checkmark & 0.8503 \\
			\Checkmark & \Checkmark & --- & \Checkmark & 0.8470 \\
			\Checkmark & \Checkmark & \Checkmark & --- & 0.8022 \\
			\hline
			\Checkmark & \Checkmark & \Checkmark & \Checkmark & \textbf{0.8849} \\
			\Xhline{1.2pt}
		\end{tabular}
	\end{center}
    \caption{The Ablation Study on MSTF}
\end{table}

\begin{figure}[t]
	\centering
	{\includegraphics[width=3.1in]{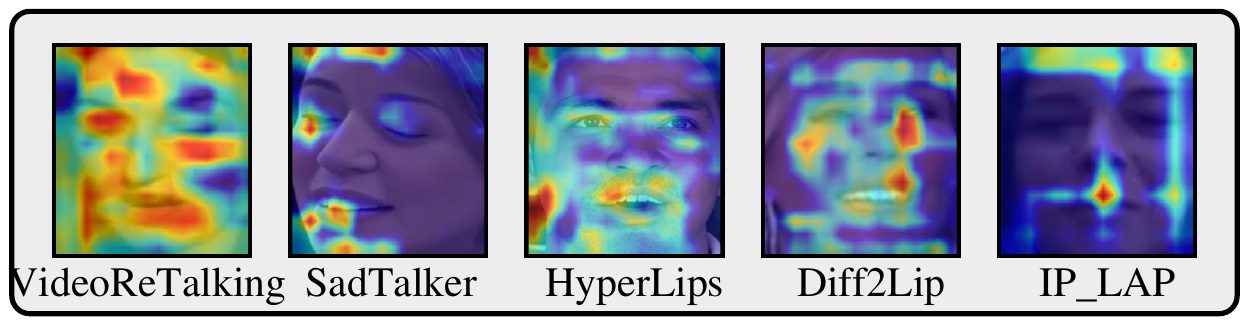}}
	\caption{Grad-CAM \cite{selvaraju2017grad} of model outputs.}
	\label{grad}
\end{figure}

\subsection{Visualization Analysis}
In Figure \ref{grad}, we draw the attention map, which shows that the model can pay attention to the forged traces in different areas, especially the mouth and eyes. It also demonstrates the effectiveness of our V-AFM for audiovisual coherence detection. The model also focuses on major motion regions such as the head and some other small regions, demonstrating the effectiveness of our RFSDM and DCTAM.

\subsection{Limitations}
Further research is needed in the future to develop detection methods specifically aimed at TFG videos compressed by social media platforms.

\section{Conclusion}
In this paper, we propose the first large-scale multi-scenario talking face dataset. For TFG detection, we propose RSFDM, which analyzes the coherence of the transition of video frames. DCTAM is introduced to bridge the process by capturing subtle inter-frame differences and adaptively aggregating spatial information. V-AFM is proposed to detect the coherence between the fused information and the corresponding audio information in the local temporal segment. Experiments demonstrate the challenge and necessity of our dataset. Our Framework also achieves optimal performance in both talking face detection and deepfake detection. 

\section{Acknowledgments}
This work is supported by the National Natural Science Foundation of China (No. U2001202, No. 62072480, No. 62172435, No. U23A20305), the National Key Research and Development Program of China (No. 2022YFB3102900), the Macau Science and Technology Development Foundation (No. SKLIOTSC-2021-2023, 0022/2022/A), the Natural Science Foundation of Guangdong Province of China (No. EF2023-00116-FST), the Open Research Project of the Key Laboratory of Artificial Intelligence Security, Ministry of Public Security (RGZNAQ-2403).

\bibliography{reference}

\begin{thebibliography}{40}
\providecommand{\natexlab}[1]{#1}

\bibitem[{Ba et~al.(2024)Ba, Liu, Liu, Wu, Lin, Lu, and Ren}]{ba2024exposing}
Ba, Z.; Liu, Q.; Liu, Z.; Wu, S.; Lin, F.; Lu, L.; and Ren, K. 2024.
\newblock Exposing the deception: Uncovering more forgery clues for deepfake detection.
\newblock In \emph{Proceedings of the AAAI Conference on Artificial Intelligence}, volume~38, 719--728.

\bibitem[{Cai et~al.(2023)Cai, Ghosh, Dhall, Gedeon, Stefanov, and Hayat}]{cai2023glitch}
Cai, Z.; Ghosh, S.; Dhall, A.; Gedeon, T.; Stefanov, K.; and Hayat, M. 2023.
\newblock Glitch in the matrix: A large scale benchmark for content driven audio--visual forgery detection and localization.
\newblock \emph{Computer Vision and Image Understanding}, 236: 103818.

\bibitem[{Cao et~al.(2014)Cao, Cooper, Keutmann, Gur, Nenkova, and Verma}]{cao2014crema}
Cao, H.; Cooper, D.~G.; Keutmann, M.~K.; Gur, R.~C.; Nenkova, A.; and Verma, R. 2014.
\newblock Crema-d: Crowd-sourced emotional multimodal actors dataset.
\newblock \emph{IEEE Transactions on Affective Computing}, 5(4): 377--390.

\bibitem[{Cao et~al.(2022)Cao, Ma, Yao, Chen, Ding, and Yang}]{cao2022end}
Cao, J.; Ma, C.; Yao, T.; Chen, S.; Ding, S.; and Yang, X. 2022.
\newblock End-to-end reconstruction-classification learning for face forgery detection.
\newblock In \emph{Proceedings of the IEEE/CVF Conference on Computer Vision and Pattern Recognition}, 4113--4122.

\bibitem[{Chen et~al.(2023)Chen, Yao, Li, Wang, Zhang, Yang, and Wen}]{chen2023hyperlips}
Chen, Y.; Yao, Y.; Li, Z.; Wang, W.; Zhang, Y.; Yang, H.; and Wen, X. 2023.
\newblock HyperLips: Hyper Control Lips with High Resolution Decoder for Talking Face Generation.
\newblock \emph{arXiv preprint arXiv:2310.05720}.

\bibitem[{Cheng et~al.(2022)Cheng, Cun, Zhang, Xia, Yin, Zhu, Wang, Wang, and Wang}]{cheng2022videoretalking}
Cheng, K.; Cun, X.; Zhang, Y.; Xia, M.; Yin, F.; Zhu, M.; Wang, X.; Wang, J.; and Wang, N. 2022.
\newblock Videoretalking: Audio-based lip synchronization for talking head video editing in the wild.
\newblock In \emph{SIGGRAPH Asia 2022 Conference Papers}, 1--9.

\bibitem[{Chung, Nagrani, and Zisserman(2018)}]{chung2018voxceleb2}
Chung, J.~S.; Nagrani, A.; and Zisserman, A. 2018.
\newblock Voxceleb2: Deep speaker recognition.
\newblock \emph{arXiv preprint arXiv:1806.05622}.

\bibitem[{Dolhansky et~al.(2020)Dolhansky, Bitton, Pflaum, Lu, Howes, Wang, and Ferrer}]{dolhansky2020deepfake}
Dolhansky, B.; Bitton, J.; Pflaum, B.; Lu, J.; Howes, R.; Wang, M.; and Ferrer, C.~C. 2020.
\newblock The deepfake detection challenge (dfdc) dataset.
\newblock \emph{arXiv preprint arXiv:2006.07397}.

\bibitem[{Feng, Chen, and Owens(2023)}]{feng2023self}
Feng, C.; Chen, Z.; and Owens, A. 2023.
\newblock Self-supervised video forensics by audio-visual anomaly detection.
\newblock In \emph{Proceedings of the IEEE/CVF Conference on Computer Vision and Pattern Recognition}, 10491--10503.

\bibitem[{Frank and Sch{\"o}nherr(2021)}]{frank2021wavefake}
Frank, J.; and Sch{\"o}nherr, L. 2021.
\newblock Wavefake: A data set to facilitate audio deepfake detection.
\newblock \emph{arXiv preprint arXiv:2111.02813}.

\bibitem[{Jiang et~al.(2020)Jiang, Li, Wu, Qian, and Loy}]{jiang2020deeperforensics}
Jiang, L.; Li, R.; Wu, W.; Qian, C.; and Loy, C.~C. 2020.
\newblock Deeperforensics-1.0: A large-scale dataset for real-world face forgery detection.
\newblock In \emph{Proceedings of the IEEE/CVF Conference on Computer Vision and Pattern Recognition}, 2889--2898.

\bibitem[{Karras et~al.(2018)Karras, Aila, Laine, and Lehtinen}]{karras2018progressive}
Karras, T.; Aila, T.; Laine, S.; and Lehtinen, J. 2018.
\newblock Progressive growing of gans for improved quality, stability, and variation. arXiv 2017.
\newblock \emph{arXiv preprint arXiv:1710.10196}, 1--26.

\bibitem[{Khalid et~al.(2021)Khalid, Tariq, Kim, and Woo}]{khalid2021fakeavceleb}
Khalid, H.; Tariq, S.; Kim, M.; and Woo, S.~S. 2021.
\newblock FakeAVCeleb: A novel audio-video multimodal deepfake dataset.
\newblock \emph{arXiv preprint arXiv:2108.05080}.

\bibitem[{Korshunov and Marcel(2018)}]{korshunov2018deepfakes}
Korshunov, P.; and Marcel, S. 2018.
\newblock Deepfakes: a new threat to face recognition.
\newblock \emph{Assessment and detection}.

\bibitem[{Kwon et~al.(2021)Kwon, You, Nam, Park, and Chae}]{kwon2021kodf}
Kwon, P.; You, J.; Nam, G.; Park, S.; and Chae, G. 2021.
\newblock Kodf: A large-scale korean deepfake detection dataset.
\newblock In \emph{Proceedings of the IEEE/CVF International Conference on Computer Vision}, 10744--10753.

\bibitem[{Ma et~al.(2022)Ma, Yi, Wang, Yan, Tao, Wang, Wang, and Fu}]{ma2022cfad}
Ma, H.; Yi, J.; Wang, C.; Yan, X.; Tao, J.; Wang, T.; Wang, S.; and Fu, R. 2022.
\newblock CFAD: A Chinese dataset for fake audio detection.
\newblock \emph{arXiv preprint arXiv:2207.12308}.

\bibitem[{Mejri, Papadopoulos, and Aouada(2021)}]{mejri2021leveraging}
Mejri, N.; Papadopoulos, K.; and Aouada, D. 2021.
\newblock Leveraging high-frequency components for deepfake detection.
\newblock In \emph{IEEE 23rd International Workshop on Multimedia Signal Processing (MMSP)}, 1--6.

\bibitem[{Mukhopadhyay et~al.(2024)Mukhopadhyay, Suri, Gadde, and Shrivastava}]{mukhopadhyay2024diff2lip}
Mukhopadhyay, S.; Suri, S.; Gadde, R.~T.; and Shrivastava, A. 2024.
\newblock Diff2lip: Audio conditioned diffusion models for lip-synchronization.
\newblock In \emph{Proceedings of the IEEE/CVF Winter Conference on Applications of Computer Vision}, 5292--5302.

\bibitem[{Nagrani, Chung, and Zisserman(2017)}]{nagrani2017voxceleb}
Nagrani, A.; Chung, J.~S.; and Zisserman, A. 2017.
\newblock Voxceleb: a large-scale speaker identification dataset.
\newblock \emph{arXiv preprint arXiv:1706.08612}.

\bibitem[{Narayan et~al.(2023)Narayan, Agarwal, Thakral, Mittal, Vatsa, and Singh}]{narayan2023df}
Narayan, K.; Agarwal, H.; Thakral, K.; Mittal, S.; Vatsa, M.; and Singh, R. 2023.
\newblock Df-platter: Multi-face heterogeneous deepfake dataset.
\newblock In \emph{Proceedings of the IEEE/CVF Conference on Computer Vision and Pattern Recognition}, 9739--9748.

\bibitem[{Panayotov et~al.(2015)Panayotov, Chen, Povey, and Khudanpur}]{panayotov2015librispeech}
Panayotov, V.; Chen, G.; Povey, D.; and Khudanpur, S. 2015.
\newblock Librispeech: an asr corpus based on public domain audio books.
\newblock In \emph{IEEE International Conference on Acoustics, Speech and Signal Processing (ICASSP)}, 5206--5210.

\bibitem[{Pang et~al.(2024)Pang, Wang, Ye, Cheung, Zhou, Huang, and Wen}]{pang2024heterogeneous}
Pang, M.; Wang, B.; Ye, M.; Cheung, Y.-M.; Zhou, Y.; Huang, W.; and Wen, B. 2024.
\newblock Heterogeneous Prototype Learning From Contaminated Faces Across Domains via Disentangling Latent Factors.
\newblock \emph{IEEE Transactions on Neural Networks and Learning Systems}.

\bibitem[{Peng et~al.(2024)Peng, Miao, Liu, Wang, Hu, and Gao}]{peng2024deepfakes}
Peng, C.; Miao, Z.; Liu, D.; Wang, N.; Hu, R.; and Gao, X. 2024.
\newblock Where Deepfakes Gaze at? Spatial-Temporal Gaze Inconsistency Analysis for Video Face Forgery Detection.
\newblock \emph{IEEE Transactions on Information Forensics and Security}.

\bibitem[{Qian et~al.(2020)Qian, Yin, Sheng, Chen, and Shao}]{qian2020thinking}
Qian, Y.; Yin, G.; Sheng, L.; Chen, Z.; and Shao, J. 2020.
\newblock Thinking in frequency: Face forgery detection by mining frequency-aware clues.
\newblock In \emph{European Conference on Computer Vision}, 86--103. Springer.

\bibitem[{Rossler et~al.(2019)Rossler, Cozzolino, Verdoliva, Riess, Thies, and Nie{\ss}ner}]{rossler2019faceforensics++}
Rossler, A.; Cozzolino, D.; Verdoliva, L.; Riess, C.; Thies, J.; and Nie{\ss}ner, M. 2019.
\newblock Faceforensics++: Learning to detect manipulated facial images.
\newblock In \emph{Proceedings of the IEEE/CVF International Conference on Computer Vision}, 1--11.

\bibitem[{Selvaraju et~al.(2017)Selvaraju, Cogswell, Das, Vedantam, Parikh, and Batra}]{selvaraju2017grad}
Selvaraju, R.~R.; Cogswell, M.; Das, A.; Vedantam, R.; Parikh, D.; and Batra, D. 2017.
\newblock Grad-cam: Visual explanations from deep networks via gradient-based localization.
\newblock In \emph{Proceedings of the IEEE International Conference on Computer Vision}, 618--626.

\bibitem[{Son~Chung et~al.(2017)Son~Chung, Senior, Vinyals, and Zisserman}]{son2017lip}
Son~Chung, J.; Senior, A.; Vinyals, O.; and Zisserman, A. 2017.
\newblock Lip reading sentences in the wild.
\newblock In \emph{Proceedings of the IEEE Conference on Computer Vision and Pattern Recognition}, 6447--6456.

\bibitem[{Sung, Chen, and Chen(2023)}]{sung2023hearing}
Sung, C.-S.; Chen, J.-C.; and Chen, C.-S. 2023.
\newblock Hearing and seeing abnormality: Self-supervised audio-visual mutual learning for deepfake detection.
\newblock In \emph{IEEE International Conference on Acoustics, Speech and Signal Processing (ICASSP)}, 1--5.

\bibitem[{Tan et~al.(2024)Tan, Zhao, Wei, Gu, Liu, and Wei}]{tan2024frequency}
Tan, C.; Zhao, Y.; Wei, S.; Gu, G.; Liu, P.; and Wei, Y. 2024.
\newblock Frequency-aware deepfake detection: Improving generalizability through frequency space learning.
\newblock \emph{arXiv preprint arXiv:2403.07240}.

\bibitem[{Todisco et~al.(2019)Todisco, Wang, Vestman, Sahidullah, Delgado, Nautsch, Yamagishi, Evans, Kinnunen, and Lee}]{todisco2019asvspoof}
Todisco, M.; Wang, X.; Vestman, V.; Sahidullah, M.; Delgado, H.; Nautsch, A.; Yamagishi, J.; Evans, N.; Kinnunen, T.; and Lee, K.~A. 2019.
\newblock ASVspoof 2019: Future horizons in spoofed and fake audio detection.
\newblock \emph{arXiv preprint arXiv:1904.05441}.

\bibitem[{Wang and Deng(2021)}]{wang2021representative}
Wang, C.; and Deng, W. 2021.
\newblock Representative forgery mining for fake face detection.
\newblock In \emph{Proceedings of the IEEE/CVF Conference on Computer Vision and Pattern Recognition}, 14923--14932.

\bibitem[{Wang et~al.(2020)Wang, Wu, Song, Yang, Wu, Qian, He, Qiao, and Loy}]{wang2020mead}
Wang, K.; Wu, Q.; Song, L.; Yang, Z.; Wu, W.; Qian, C.; He, R.; Qiao, Y.; and Loy, C.~C. 2020.
\newblock Mead: A large-scale audio-visual dataset for emotional talking-face generation.
\newblock In \emph{European Conference on Computer Vision}, 700--717. Springer.

\bibitem[{Wang et~al.(2024)Wang, Ye, Tang, Zhang, and Deng}]{wang2024avt2}
Wang, R.; Ye, D.; Tang, L.; Zhang, Y.; and Deng, J. 2024.
\newblock AVT2-DWF: Improving Deepfake Detection with Audio-Visual Fusion and Dynamic Weighting Strategies.
\newblock \emph{arXiv preprint arXiv:2403.14974}.

\bibitem[{Yang et~al.(2023)Yang, Zhou, Chen, Guo, Ba, Xia, Cao, and Ren}]{yang2023avoid}
Yang, W.; Zhou, X.; Chen, Z.; Guo, B.; Ba, Z.; Xia, Z.; Cao, X.; and Ren, K. 2023.
\newblock Avoid-df: Audio-visual joint learning for detecting deepfake.
\newblock \emph{IEEE Transactions on Information Forensics and Security}, 18: 2015--2029.

\bibitem[{Yang et~al.(2024)Yang, Zeng, Guo, Wang, Dong, and Wang}]{yang2024robust}
Yang, X.; Zeng, J.; Guo, D.; Wang, S.; Dong, J.; and Wang, M. 2024.
\newblock Robust Video Question Answering via Contrastive Cross-Modality Representation Learning.
\newblock \emph{SCIENCE CHINA Information Sciences}, 67: 1--16.

\bibitem[{Zhang et~al.(2023)Zhang, Cun, Wang, Zhang, Shen, Guo, Shan, and Wang}]{zhang2023sadtalker}
Zhang, W.; Cun, X.; Wang, X.; Zhang, Y.; Shen, X.; Guo, Y.; Shan, Y.; and Wang, F. 2023.
\newblock Sadtalker: Learning realistic 3d motion coefficients for stylized audio-driven single image talking face animation.
\newblock In \emph{Proceedings of the IEEE/CVF Conference on Computer Vision and Pattern Recognition}, 8652--8661.

\bibitem[{Zhao et~al.(2021)Zhao, Zhou, Chen, Wei, Zhang, and Yu}]{zhao2021multi}
Zhao, H.; Zhou, W.; Chen, D.; Wei, T.; Zhang, W.; and Yu, N. 2021.
\newblock Multi-attentional deepfake detection.
\newblock In \emph{Proceedings of the IEEE/CVF Conference on Computer Vision and Pattern Recognition}, 2185--2194.

\bibitem[{Zhong et~al.(2023)Zhong, Fang, Cai, Wei, Zhao, Lin, and Li}]{zhong2023identity}
Zhong, W.; Fang, C.; Cai, Y.; Wei, P.; Zhao, G.; Lin, L.; and Li, G. 2023.
\newblock Identity-preserving talking face generation with landmark and appearance priors.
\newblock In \emph{Proceedings of the IEEE/CVF Conference on Computer Vision and Pattern Recognition}, 9729--9738.

\bibitem[{Zhou et~al.(2021)Zhou, Wang, Liang, and Shen}]{zhou2021face}
Zhou, T.; Wang, W.; Liang, Z.; and Shen, J. 2021.
\newblock Face forensics in the wild.
\newblock In \emph{Proceedings of the IEEE/CVF Conference on Computer Vision and Pattern Recognition}, 5778--5788.

\bibitem[{Zhou and Lim(2021)}]{zhou2021joint}
Zhou, Y.; and Lim, S.-N. 2021.
\newblock Joint audio-visual deepfake detection.
\newblock In \emph{Proceedings of the IEEE/CVF International Conference on Computer Vision}, 14800--14809.

\end{thebibliography}

\end{document}